\documentclass{article}

\usepackage{microtype}
\usepackage{graphicx}
\usepackage{subfigure}
\usepackage{booktabs} 
\usepackage{multirow}
\usepackage{array}
\usepackage{longtable}     
\usepackage{makecell}

\usepackage{hyperref}

\usepackage[accepted]{icml2025}

\usepackage{amsmath}
\usepackage{amssymb}
\usepackage{mathtools}
\usepackage{amsthm}

\usepackage[capitalize,noabbrev]{cleveref}

\theoremstyle{plain}

\theoremstyle{definition}

\theoremstyle{remark}

\usepackage[textsize=tiny]{todonotes}

\icmltitlerunning{Technical AI Governance Workshop -- ICML 2025}

\begin{document}

\twocolumn[
\icmltitle{A Conceptual Framework for AI Capability Evaluations}



\icmlsetsymbol{equal}{*}

\begin{icmlauthorlist}
\icmlauthor{María Victoria Carro}{gen,fair}
\icmlauthor{Denise Alejandra Mester}{fair}
\icmlauthor{Francisca Gauna Selasco}{fair}
\icmlauthor{Luca Nicolás Forziati Gangi}{fair}
\icmlauthor{Matheo Sandleris Musa}{uba}
\icmlauthor{Lola Ramos Pereyra}{fair}
\icmlauthor{Mario Leiva}{uns,icic}
\icmlauthor{Juan Gustavo Corvalan}{uba}
\icmlauthor{Maria Vanina Martinez}{iiia}
\icmlauthor{Gerardo Simari}{uns,icic,asu}
\end{icmlauthorlist}

\icmlaffiliation{gen}{Università degli Studi di Genova, GE, Italy}
\icmlaffiliation{uba}{University of Buenos Aires, BA, Argentina}
\icmlaffiliation{fair}{FAIR, IALAB, University of Buenos Aires, BA, Argentina}
\icmlaffiliation{uns}{Dept. of Computer Science and Engineering, Universidad Nacional del Sur (UNS)}
\icmlaffiliation{icic}{Inst. of Computer Science and Engineering (ICIC UNS-CONICET), Bahía Blanca, BA, Argentina}
\icmlaffiliation{iiia}{Artificial Intelligence Research Institute (IIIA-CSIC), Universidad Autónoma de Barcelona, Barcelona, España}
\icmlaffiliation{asu}{School of Computing and Augmented Intelligence, Arizona State University, USA}

\icmlcorrespondingauthor{María Victoria Carro}{6381013@studenti.unige.it}

\icmlkeywords{Machine Learning, ICML}

\vskip 0.3in
]



\printAffiliationsAndNotice{} 

\begin{abstract}
As AI systems advance and integrate into society, well-designed and transparent evaluations are becoming essential tools in AI governance, informing decisions by providing evidence about system capabilities and risks. Yet there remains a lack of clarity on how to perform these assessments both comprehensively and reliably. To address this gap, we propose a conceptual framework for analyzing AI capability evaluations, offering a structured, descriptive approach that systematizes the analysis of widely used methods and terminology without imposing new taxonomies or rigid formats. This framework supports transparency, comparability, and interpretability across diverse evaluations. It also enables researchers to identify methodological weaknesses, assists practitioners in designing evaluations, and provides policymakers with an accessible tool to scrutinize, compare, and navigate complex evaluation landscapes.
\end{abstract}

\section{Introduction}
Evaluations are gaining significant attention in the field of AI, with substantial efforts dedicated to advancing this area. The rapid evolution of large language models (LLMs) and their growing integration into daily life have underscored the need for robust and rigorous processes to understand the state of frontier AI current capabilities, identify potential risks to improve safety, and comprehend its societal impact. 

In particular, capability evaluations\footnote{Capability evaluations have been defined as those that comprehensively assess a system’s overall capabilities, including planned, unplanned, emerging, or dangerous capabilities \cite{xia2024ai}. \citet{burden2025paradigms} note that capability evaluations are used by AI developers and regulators to determine whether a system is safe to deploy, and by AI adopters to assess whether a system can automate specific tasks within their organizations. In this paper, we adopt a broad definition of capability evaluations, aligned with the usage found in \citet{10.5555/3692070.3693800}. Accordingly, we exclude impact evaluations—also referred to as “real-world impact evaluations”\cite{burden2025paradigms}—which measure the effects of AI systems once deployed in real-world settings. These evaluations typically involve human subjects and treat the AI system’s assistance as an “intervention,” whose effect must be empirically quantified.} are crucial for various stakeholders, including academia, industry, government, and end-users. They serve as essential tools for tracking and communicating progress within the AI community, and for demonstrating the improvements of newly proposed methods over prior baselines \cite{biderman2024lessons}.  Additionally, they play a key role in defining the threshold for achieving artificial general intelligence (AGI) \cite{pfister2025understanding, bubeck2023sparks}.

Evaluations are key components of governance regimes \cite{reuel2025openproblemstechnicalai, mokander2023auditing}. In regulatory contexts, they help to identify areas where intervention may be needed and inform decisions by providing evidence about system capabilities and risks \cite{reuel2025openproblemstechnicalai, shevlane2023model, paskov2025preliminary}. Notably, the EU AI Act incorporates benchmarks in several key provisions \cite{eriksson2025can}, becoming the world’s first mover in government-mandated general-purpose AI (GPAI) evaluations \cite{paskovgpai}. 

For evaluations to fulfill these roles effectively, they must be conducted and reported rigorously \cite{paskov2025preliminary}, with sufficient context and transparency to allow for meaningful interpretation by a wide range of stakeholders, including policymakers \cite{staufer2025audit}. However, there is a lack of clarity on how to perform these assessments both comprehensively and reliably \cite{10.5555/3692070.3693800}, and existing practices are among the key factors shaping practitioners' evaluation choices \cite{zhou2022deconstructing}, which may not align with the informational needs of decision-makers. Without well-designed evaluations, governments risk relying on incomplete, misleading, or selectively reported information, undermining efforts to ensure the safe and beneficial development and deployment of AI systems.

In this paper, we propose a conceptual framework for analysing capability evaluations, which includes key elements and sub-elements of evaluation processes, their interrelationships, and examples with relevant variables. Additionally, some of the key challenges associated with evaluations are identified and integrated into the framework in the extended version of \hyperref[appendix: 1]{Appendix 1}. By creating a highly simplified representation of complex processes, essential features can be abstracted to permit systematic reasoning \cite{gupta2024conceptual}. This, in turn, provides a clear overview and a structured approach that supports the standardization of concepts, methods and reports—recognized as a critical need in the field \cite{weidinger2023sociotechnical, thurnherr2024who}—, as well as the comparison and transparency of evaluation processes.  Such transparency is especially important when evaluations are developed by independent third parties, as policymakers and other stakeholders may need to scrutinize the results, something that depends, in part, on how well the evaluation’s development and execution are documented \cite{thurnherr2024who}.

We developed this framework through a careful analysis of the design of numerous evaluations, some of which are cited as examples throughout the paper. While not exhaustive, it is intended to be comprehensive—capturing the full range of aspects that an evaluation method may encompass, even though not all elements will be present in every individual evaluation (e.g., prompt techniques).

Our conceptual framework offers two main advantages. First, it adopts existing terminology and systematizes current methods, integrating the most widely used concepts in the field rather than introducing new terms. In doing so, it adapts to the established language of diverse stakeholders, rather than requiring them to conform to a novel taxonomy. The framework is thus descriptive rather than normative. However, in instances where recommendations or best practices are mentioned, these are drawn from the existing literature and included based on their broad acceptance within the field, rather than originating from the authors themselves. Second, it prioritizes simplicity and accessibility, avoiding unnecessary complexity or rigid formats. It aims to foster consensus around a broad, intuitive framework that remains easily understandable to all participants in the field of evaluations, including those without deep technical expertise.

This approach has the potential to support a wide range of stakeholders in navigating and interpreting AI capability evaluations: 
\begin{itemize}
\item For \textbf{public bodies}, it offers a structured tool to scrutinize, audit and systematically compare evaluations, enabling more consistent assessments. 

\item For \textbf{researchers} and \textbf{third-party evaluators}, it provides a conceptual map to understand which dimensions matter for use cases, synthesize and critically analyze evaluation practices, identify methodological strengths, weaknesses, and gaps.

\item For \textbf{industry practitioners} and \textbf{model developers}—who must make numerous decisions throughout the evaluation process \cite{gupta2024conceptual}—this highlights key elements to consider when designing or reporting evaluations, promoting greater transparency and comparability. 

\item For \textbf{newcomers to the field} and \textbf{policymakers}, it serves as an accessible entry point to understand evaluation structures and interpret results. 
\end{itemize}
\label{sec:introduction}

\section{Related Work}
\citet{dow2024dimensions} present Dimensions of Generative AI Evaluation Design for mapping the state of evaluations in which many proposed dimensions—such as input source, task type, and metric type—overlap with elements of our own. However, their framework is limited to generative AI systems and safety evaluations, whereas ours may be applicable more broadly. While the underlying motivation is similar—highlighting the importance of structured evaluation—our framework goes further by decomposing the elements into subcomponents, systematically integrating evaluation challenges within each element, and providing a more granular and detailed perspective. 

Additional related work includes \citet{laskar2024systematic}, who survey the challenges of evaluation and structure their analysis by segmenting the evaluation process into distinct stages. Furthermore, \citet{paskov2025preliminary} outline suggestions for enhancing the rigor of general-purpose AI (GPAI) evaluations, offering practical recommendations for each stage of the evaluation life cycle. While the latter primarily focuses on benchmarks and uplift studies, several of their insights—such as documentation practices and statistical safeguards—have broader applicability. However, both works rely on a sequential model of the evaluation process, prescribing a particular order of steps. In contrast, our framework adopts a descriptive approach: rather than prescribing what should happen first or next, it maps the key elements of evaluation without imposing a fixed sequence.

Finally, while \citet{weidinger2023sociotechnical} propose and construct a repository of existing evaluations, their work is limited to risk evaluations, which are categorized according to a risk taxonomy introduced by the authors. Moreover, the repository is a static resource, though they suggest the development of a living version as future work. In contrast, we propose a broader repository that includes, but is not limited to, risk evaluations, and organizes them according to multiple criteria. This repository is envisioned as an interactive resource—similar in spirit to the MIT Risk Repository \cite{slattery2024ai}, but focused specifically on evaluations.

\section{The Conceptual Framework}

\subsection{Evaluation Target}

\textbf{Capability.} Capability evaluations focus on what AI systems can and cannot do. The specific capability under evaluation might range from mathematical reasoning \cite{didolkar2024metacognitive} and causal inference \cite{kiciman2023causal} to legal reasoning \cite{guha2023legalbench}, deception \cite{hagendorff2024deception}, and many others.

\textbf{Objective.} The objective of the evaluation may be to quantify system capabilities, track progress, enable large-scale comparisons, understand behavioral patterns, identify and estimate potential risks \cite{weidinger2023sociotechnical}, provide assurance of system safety \cite{burden2025paradigms, weidinger2023sociotechnical}, predict if future models can lead to catastrophic harm \cite{barnett2024declare} or evaluate evaluation methods, among others.

\subsection{Task}
The \textbf{task} refers to a particular specification of a problem \cite{raji2021ai}, typically represented as a mapping between an input space and an output or action space \cite{schlangen2021targeting, eriksson2025can}. 

\begin{figure}[t]
  \centering
  \includegraphics[width=\linewidth]{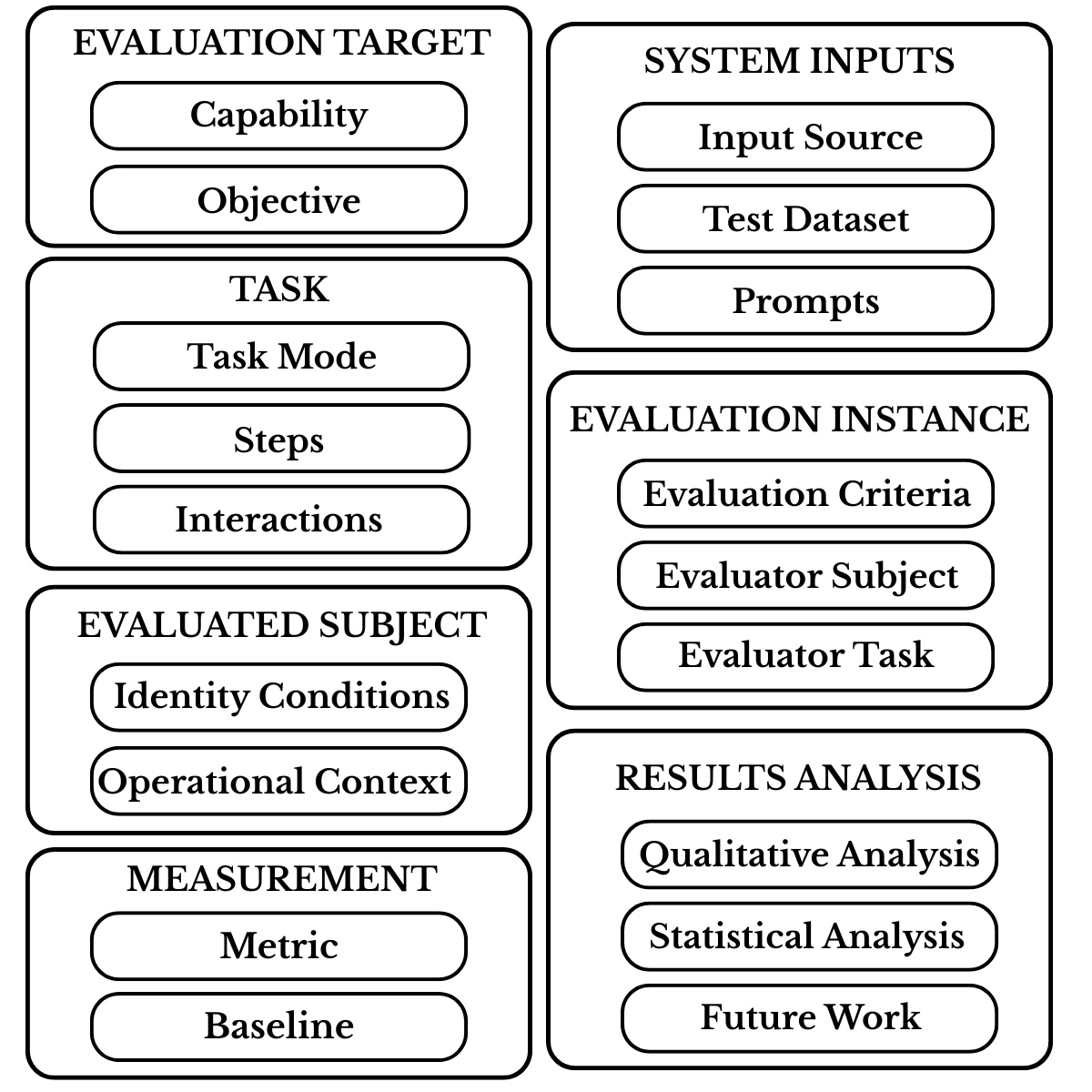}
  \caption{Overview of the proposed Conceptual Framework.}
  \label{fig:image1}
\end{figure}

\textbf{Task Mode.} The task mode is the format of the response required for the system, which can be closed-ended, when the subject is asked to identify an answer from a predetermined set of options, or open-ended, when the subject is required to generate a novel output \cite{burden2025paradigms}. Some examples include question answering, text completion, text summarization, or error identification and correction. 

\textbf{Steps.} Some evaluations require the system to perform multiple coordinated steps to complete a task, to simulate real-world activities \cite{kraprayoon2025ai}. These intermediate steps can be autonomously defined by the system or dictated by a human subject. This is particularly relevant for AI agent evaluations, which may involve end-to-end processes that include planning, tool use, and iterative refinement to produce a coherent final output \cite{testini2025measuring}.

\textbf{Interactions.} Number of interactions between the evaluated subject and the user or evaluator, which can be classified as single-turn or multi-turn \cite{wang2024mint, cheng2024leveraging, sirdeshmukh2025multichallenge}. 

\subsection{Evaluated Subject}
The \textbf{evaluated subject} is the subject of the capability claim—in most cases, an AI system, and in the case of meta-evaluations, an evaluation method itself (e.g. \citet{hong2024curiositydriven}). 

\textbf{Identity Conditions.} The evaluated system should be individualized by defining its identity conditions \cite{harding2024machine, biderman2024lessons}. This includes, where possible, fixed attributes such as details of the architecture and parameters \cite{paskov2025preliminary}, the system’s name and version, and whether it is a foundational model or a fine-tuned variant.

\textbf{Operational Context.} The operational context is the background conditions of the evaluation \cite{harding2024machine}, including access methods (e.g., raw API or assistant interface), configuration settings (e.g. temperature), auxiliary tools (e.g. plugins), inference budget (e.g. token or query limits), and hardware or compute conditions where applicable. 

\subsection{System Inputs}
The \textbf{input source} refers to the origin of the inputs used in the evaluation \cite{dow2024dimensions}. These may come from a pre-existing test dataset, or user-generated inputs, including those produced dynamically interacting with the system.  For example, in red teaming, attacks on the target model are often conducted through trial and error to identify effective strategies that elicit abnormal behavior \cite{lin2025against}. The input can also take various modalities, such as image, audio, video, text, or multimodal combinations.

\textbf{Test Dataset.} A separate dataset, distinct from the training data, consisting of data instances can be used to evaluate the AI system on a given task. However, in contexts where generalization is not prioritized, the test data may overlap with the training data. In the case of benchmarks and reference-based evaluations, the dataset also includes the desired output and annotations for each instance \cite{eriksson2025can, liu2024datasets}. Key dimensions to consider include the dataset’s size and construction method—for example, whether it was manually developed by researchers or domain experts, or synthetically generated, for instance, through an LLM.

\textbf{Prompt Techniques.} A prompting strategy serves to turn each data instance into an input which can be processed by an AI system. In this context, most prompting strategies can be viewed as ‘templates’, which generate the model input by embedding the query in a pre-written prompt format \cite{harding2024machine}. Furthermore, prompts may include model-specific instructions designed to steer behavior and optimize performance \cite{DBLP:journals/corr/abs-2406-06608}. Some of the prompt techniques include for example, few shots, one shot, chain-of-thought (CoT), self-consistency or emotion prompting \cite{DBLP:journals/corr/abs-2402-07927, DBLP:journals/corr/abs-2406-06608}.

\subsection{Evaluation Instance}
\textbf{Evaluation Criteria.} This refers to specific aspects of the output that define what constitutes a correct, appropriate, or high-quality response. Some examples are correctness, fluency, generalization, reasoning, robustness \cite{hu2024unveiling}, or logical coherence.

\textbf{Evaluator Subject.} An evaluator subject—human or AI system—is needed when the evaluation criteria are not mathematically defined and no prior annotation of the dataset exists. It is key to consider how the evaluator is selected—e.g., human evaluators may be recruited via crowdsourcing, while, ideally, LLM evaluators should not be the same model under evaluation or from the same model family, to avoid self-preference bias \cite{panickssery2024llm, bai2023benchmarking}.

\textbf{Evaluator Task.} The nature and instructions of the evaluator’s task must be specified, such as pairwise comparison, as in Chatbot Arena \cite{chiang2024chatbot}, rating individual outputs on a Likert scale or assigning a score. The order in which outputs are presented is also important to avoid position bias \cite{van2019best, gao2025llm}.

Note that in this framework, we distinguish between the concepts of annotators and evaluators. An \textit{annotator}—whether a human or an AI system—is responsible for labeling a test dataset prior to the evaluation of the evaluated subject. This may involve selecting the correct answer in a multiple-choice item, assigning categories to text fragments, or specifying the desired answer to a question. In contrast, an \textit{evaluator} assesses and annotates the outputs of the evaluated subject, for instance by rating the creativity of a poem generated by an LLM using a Likert scale. While the term `annotator' is often used in the literature to refer to both roles (e.g. \citet{weidinger2023sociotechnical}), we maintain this distinction for clarity.

\subsection{Measurement}
\textbf{Metric.} A metric is a mathematically defined function used to quantify a specific evaluation criterion. Metrics provide a standardized and objective way to measure aspects of system outputs, as they can summarize model performance on a set of tasks and datasets as a single number or score \cite{raji2021ai, eriksson2025can}. They can be classified as reference-based or reference-free \cite{ito2025reference, li2024leveraging}, and as general-purpose (e.g. Accuracy, Recall, Precision) or task-specific metrics (e.g. Translation Edit Rate (TER) \cite{snover2006study}).

Evaluation practitioners may confuse evaluation criteria with the metrics used to operationalize them, but this confusion risks losing the ability to assess the metric’s appropriateness and distinguish between the metric and the original goal \cite{zhou2022deconstructing}. As Goodhart’s Law states, `When a measure becomes a target, it ceases to be a good measure'.

\textbf{Baselines.} Baselines are typically used as reference points or standards for comparison when the goal of an evaluation is to compare the performance of different systems, approaches, or evaluation methods.

\subsection{Result Analysis}
\textbf{Qualitative Analysis.} Ranking systems according to a single quality number is easy and actionable, but often, it is much more important to understand when and why models fail \cite{eriksson2025can}. Therefore, it is also beneficial to conduct a deeper analysis of the evaluation results (e.g., analyzing results in relation to dataset complexity levels, if a dataset composition analysis was conducted).

\textbf{Statistical Analysis.} When applicable, statistical analysis should be performed to distinguish genuine performance improvements from random variability. Best practices emphasize treating evaluation processes as structured experiments, applying significance testing and quantifying uncertainty through confidence intervals or resampling techniques \cite{miller2024adding, DBLP:journals/corr/abs-2403-15250, zhan_evaluatology_2024}.

\textbf{Future Work.} Results can reveal valuable insights and guide future research directions \cite{hamalainen-alnajjar-2021-human} as well as inform system development and highlight specific avenues for improvement.

\label{sec: the conceptual framework}

\section{Future Work and Conclusion}

In this paper, we present a conceptual framework designed to map and structure capability evaluations of AI systems. A potential direction for future work is to analyze a range of evaluations using this framework as a set of use cases and to iteratively refine it. To demonstrate its practical application, we conducted such an analysis on three papers, presented in \hyperref[appendix: 2]{Appendix 2}. Given the exponential growth in the number of evaluations being conducted and published, it is becoming increasingly difficult for stakeholders to stay up to date. This framework could serve as the basis for an interactive and dynamic repository of evaluations, enabling users to track trends in the field across different elements and challenges. For example, a researcher could explore all proposed solutions to data contamination or consult a catalog of metrics to identify those best suited to a particular evaluation. While this approach could contribute to a degree of standardization that has been recognized as necessary in the field \cite{weidinger2023sociotechnical}, it would also preserve flexibility by making relevant and up-to-date information more accessible. Ultimately, this would facilitate cumulative knowledge-building and allow the community to make more efficient progress on the field’s open challenges.
\label{sec: future work and conclusion}

\bibliography{icml2025}
\bibliographystyle{icml2025}

\newpage
\appendix
\onecolumn
\section{Appendix 1: The Conceptual Framework for Capability Evaluations}
\label{appendix: 1}

In this expanded version of the framework, several well-known challenges and limitations of evaluations are integrated within the elements and sub-elements; additional examples of specific practices or methodologies and their applications are included; and aspects of the sub-elements that the relevant literature identifies as reporting best practices are highlighted.
\subsection{Evaluation Target}
Capability evaluations focus on what AI systems can and cannot do. The specific capability under evaluation might range from mathematical reasoning (e.g.\citet{didolkar2024metacognitive, imani-etal-2023-mathprompter}) and causal inference (e.g. \citet{kiciman2023causal, jin2024can}) to legal reasoning (e.g. \citet{guha2023legalbench, bhambhoria2024evaluating}), deception (e.g. \citet{hagendorff2024deception}), and many others.

A philosophical account defines a machine learning (ML) system as having a capability to X just when it would reliably succeed at doing X if it ‘tried’ \cite{harding2024machine, burden2024evaluating}. A further useful distinction is between competence and performance: while the former refers to the possession of a capability, performance refers to its successful manifestation. Thus, a failure in performance does not necessarily imply the absence of the underlying competence \cite{harding2024machine, hagendorff2024machinepsychology}.

The objective of the evaluation may be to quantify system capabilities, track progress, enable large-scale comparisons -as in traditional benchmarks \cite{reuel2024betterbench}, understand behavioral patterns, identify and estimate potential risks, provide assurance of system safety \cite{burden2025paradigms}, predict future behavior \cite{burden2024evaluating} e.g. if upcoming models can lead to catastrophic harm \cite{barnett2024declare} or evaluate evaluation methods, among others.

An evaluation may therefore aim to compare the performance of different LLMs on document classification, or to assess a system’s capacity to generate toxic content in order to identify potential misuse risks.

It is also important to recognize that the interpretation and implications of an evaluation may depend on the specific capability or objective, as well as on the context in which the system is implemented or intended to be implemented. For instance, the same evaluation objective—such as assessing a system’s classification capability—may carry different significance depending on whether the system is deployed in a critical domain (e.g., healthcare, criminal justice) or a non-critical setting (e.g., entertainment recommendations) \cite{oecd2022oecd, thurnherr2024who}.

\subsection{Task}

As in human evaluations, there are many scenarios and tasks for evaluating a given capability. The task refers to a particular specification of a problem \cite{raji2021ai}, typically represented as a mapping between an input space and an output or action space \cite{schlangen2021targeting, eriksson2025can}.

Part of the task specification involves the format of the response required from the model, which  \citet{burden2025paradigms} called the task mode, explaining that in some cases the subject is asked to identify an answer from a predetermined set of options (e.g., selecting a multiple-choice answer or a class label) or to generate a novel output (e.g., a numerical value in a continuous range or free-form text). This distinction between closed-ended and open-ended generation tasks has a direct impact on the design of the measurement component and the evaluation instance, as will be examined in section A5. Further examples of task configurations include question answering, text completion, text summarization, debate or error identification and correction, among others. 

Furthermore, particularly in the case of agent evaluations, the task may involve performing multiple coordinated steps to accomplish complex objectives, simulating real-world scenarios \cite{kraprayoon2025ai}. These intermediate sub-tasks can be autonomously defined by the system or dictated by a human subject. For instance, when evaluating agents in data science settings, some studies assess performance on end-to-end tasks that involve planning, generating code and plots, and producing coherent outputs and insights \cite{testini2025measuring}.

The rationale for selecting the task should be described, explaining their relevance to the evaluation objective and the specific capabilities being assessed \cite{cao2025buildbenchmarkrevisiting274}. This demonstrates that the tasks are not arbitrary, but rather carefully chosen to enhance transparency and enable public scrutiny \cite{reuel2024betterbench, staufer2025audit, liang2023holistic}.

One dimension to consider is the number of interactions between the evaluated subject and the user or evaluator. Along this dimension, interactions can be classified as single-turn—as in most traditional benchmarks—or multi-turn, when they involve multiple exchanges \cite{wang2024mint, cheng2024leveraging, sirdeshmukh2025multichallenge}. A typical example of multi-turn interactions is found in evaluation methods based on debate (e.g. \citet{moniri-etal-2025-evaluating}).

The Second Draft of the GPAI Code of Practice \cite{gpaicode2024} highlights two types of validity. External validity refers to the extent to which evaluation results can be used as a proxy for model behavior in contexts outside of the evaluation environment \cite{weidinger2023sociotechnical, paskovgpai, biderman2024lessons, reuel2025openproblemstechnicalai, burden2024evaluating}. This aligns with what \citet{burden2024evaluating} describes as situational external validity, which he distinguishes from external validity across subjects—the question of whether pre-existing tests remain valid when applied to AI systems. Internal validity, on the other hand, refers to the extent to which the observed evaluation results represent the truth in the evaluation setting and are not due to methodological shortcomings \cite{weidinger2023sociotechnical, paskovgpai, reuel2025openproblemstechnicalai}. 

These principles have a direct impact on task selection. For instance, multiple-choice questions offer a fast and scalable method for evaluating model capabilities, as their closed-ended nature makes it straightforward to compute correctness. However, their external validity may be limited, as they rarely mirror the interactive, open-ended, or multimodal nature of real-world AI applications \cite{yang2023a, aisi2024qa}, as well as their dynamic characteristics \cite{mcintosh2024inadequacies, eriksson2025can}. Additionally, as \citet{bieger2016evaluation} point out, while focusing solely on final outputs—such as a completed multiple-choice test—can simplify black-box comparisons, it also risks discarding rich information about the system’s behavior and reasoning processes over time.

While these principles resonate in many critical works on benchmarks and evaluations \cite{biderman2024lessons, reuel2024betterbench, weidinger2025toward, cao2025buildbenchmarkrevisiting274, liao2023rethinking}, they may not be absolute and often depend on the specific goals of the evaluation. For instance, there are entire evaluation paradigms that do not prioritize real-world representativeness. As \citet{burden2025paradigms} describe, the Construct-Oriented Paradigm carefully designs tasks, controlling for confounding factors and often adapting tasks from cognitive science literature. Conversely, \citet{narayanan2023gpt4} critique the use of professional benchmarks by OpenAI to evaluate GPT-4, noting that \textit{'it’s not like a lawyer’s job is to answer bar exam questions all day'} \cite{eriksson2025can}. These design decisions are informed by well-established and accepted methodologies from other disciplines. While such methods may not be the most appropriately representative of real-world performance, they prioritize different foundational goals. Therefore, the field of AI evaluation should be open to considering these diverse approaches.

\subsection{Evaluated Subject}

The evaluated subject is the subject of the capability claim—in most cases, an AI system. Depending on the objective of the evaluation, this may involve a single system or multiple systems, particularly when the goal is to compare the performance of different systems with respect to the capability under evaluation. However, in the case of meta-evaluations, the subject of evaluation is an evaluation method itself —such as its design, reliability, or alignment with its intended purpose (e.g.  \citet{lee2025learning, hong2024curiositydriven, Zhang_Zhang_Yuan_Liu_Shi_Gui_Zhang_Huang_2024}).

An important consideration is to individuate the system—that is, to define its identity conditions \cite{harding2024machine}. This is essential for ensuring transparency and enabling fair comparisons—or “apples to apples”—across systems \cite{biderman2024lessons}. Although models are often characterized by their architecture and parameters \cite{harding2024machine, paskov2025preliminary}, this information is not always available in practice\footnote{The level of access granted to the evaluation practitioner shapes the range of information available. Access can take the form of black-box (limited to inputs and outputs), grey-box (partial internal insights), white-box (full access to weights, gradients, and activations), or, as introduced by \citet{casper2024black}, outside-the-box (contextual details such as training data, source code, and documentation).}. Therefore, where possible, fixed attributes that help individuate the system should be reported, including its name and version, whether it is open- or closed-source, and whether it is a foundational model or a fine-tuned variant \cite{harding2024machine}. 

Because evaluations can be highly sensitive to seemingly minor implementation details—such as interface design, prompting techniques,  access method, and whether elicitation techniques were employed to probe latent capabilities \cite{hofstatter2024the, paskov2025preliminary}—that may significantly influence model behavior \cite{biderman2024lessons, mcintosh2024inadequacies, liang2023holistic}, it is also important to report the surrounding scaffolding and operational context \cite{harding2024machine}. This includes documenting background conditions that may vary across evaluations, such as how the system was accessed (e.g., via a raw API or an assistant interface), configuration settings (e.g., temperature, context window, random seeds), any auxiliary tools or capabilities (e.g., plugins, memory, or browsing), and the inference budget, including the compute or hardware conditions under which the system was executed, where relevant \cite{harding2024machine, staufer2025audit, paskov2025preliminary}.

Additionally, it is important to consider and report intra-model variance, the variation in performance that can arise from repeated evaluations of the same model under different conditions (e.g. sampling temperatures and random seeds, as mentioned above) \cite{reuel2024betterbench}. Reporting this, can help distinguish genuine performance differences between systems from noise introduced by the evaluation \cite{reuel2024betterbench}.

\subsection{System Inputs}

The input source refers to the origin of the inputs used in the evaluation \cite{dow2024dimensions}. These may come from a pre-existing test dataset, or user-generated inputs, including those produced dynamically interacting with the system. For example, in red teaming, attacks on the target model are often conducted through trial and error to identify effective strategies that elicit abnormal behavior \cite{lin2025against}. The input can also take various modalities, such as image, audio, video, text, or multimodal combinations. 

The test dataset refers to a separate dataset, distinct from the training data, consisting of data instances used to evaluate the AI system on a given task. In other cases, when generalization is not a primary concern, such separation between test dataset and training data may not be required. In the case of benchmarks and reference-based evaluations, the dataset also includes the desired output and annotations for each instance \cite{eriksson2025can, liu2024datasets}.

The dataset may originate from widely used benchmarks or alternative sources and can undergo preparation and curation processes prior to evaluation \cite{cao2025buildbenchmarkrevisiting274}. Key dimensions to consider include the dataset’s size and construction method—for example, whether it was manually developed by researchers or domain experts, or synthetically generated by an LLM—and, where applicable, considerations related to informed consent, permissions \cite{piktus2023roots} and license compliance. Transparent reporting of these factors enables a better understanding of the dataset’s context and limitations, and supports the consideration of relevant ethical implications \cite{reuel2024betterbench, staufer2025audit}.

\citet{zhang2024gemf} argue that a good evaluation should include a mix of item difficulties. Transparency is also enhanced by analyzing the composition of the dataset in terms of task difficulty and task type, among other aspects. Often, such analysis depends on the specific capability being evaluated or the domain to which it belongs. For example, \citet{bai2023benchmarking} use Bloom’s Taxonomy to classify the questions in their dataset and calculate the distribution of question forms based on interrogative words (e.g., how, what, why). Similarly, in the field of causal reasoning, researchers have applied the Ladder of Causation to structure tasks according to levels of causal difficulty \cite{10.5555/3666122.3667475}.  Beyond domain-specific frameworks, there have also been efforts to estimate task difficulty a priori using textual features or model-based uncertainty \cite{benedetto2023quantitative, kuhn2023semantic}. To support more principled test construction, Item Response Theory (IRT) has been applied in NLP to describe characteristics of individual items—their difficulty and discriminating power—and to account for these characteristics in the estimation of system performance \cite{lalor2016building}. This approach has been used to identify datasets that are best suited for distinguishing among state-of-the-art models, as well as those that are largely saturated and unlikely to detect further progress \cite{vania2021comparing}.

According to \citet{harding2024machine} the dataset can be thought of as a particular set of queries. A prompting strategy then serves to turn each query into an input which can be processed by an LLM. In this context, most prompting strategies can be viewed as ‘templates’, which generate the model input by embedding the query in a pre-written prompt format \cite{harding2024machine}. In addition, prompts may also include model-specific instructions designed to steer the model’s behavior toward the desired outcome \cite{DBLP:journals/corr/abs-2406-06608}. Some of the prompt techniques include for example, few shots, one shot, chain-of-thought (CoT), self-consistency or emotion prompting \cite{DBLP:journals/corr/abs-2402-07927, DBLP:journals/corr/abs-2406-06608}. Since even minor variations in prompts can significantly affect model performance \cite{perez2023discovering, sclar2024quantifying, frontier2024bestpractices, mcintosh2024inadequacies, liang2023holistic}, it is important to report the full prompts used \cite{reuel2024betterbench} in order to ensure that all data is available to support the reproducibility of evaluations \cite{biderman2024lessons, cao2025buildbenchmarkrevisiting274}.

In general, the greater the number of inputs used in an evaluation, the more reliable and robust the results will be as evidence for claims about a model’s capabilities \cite{harding2024machine}. Additionally, because of the nondeterministic nature of LLMs, experiments should be repeated, and randomization strategies should be used to mitigate the effects of randomness and parameter configuration biases \cite{cao2025buildbenchmarkrevisiting274, liang2023holistic}.

An important challenge in this dimension is data contamination -also known as data leaking \cite{zhou2025lessleak}, which occurs when the test dataset overlaps with the training data \cite{mcintosh2024inadequacies}, leading to an overestimation of the system’s performance \cite{sainz2023nlp, zhou2023don, laskar2024systematic}. Numerous strategies have been proposed to identify and mitigate this issue, making evaluations ``google-proof'' \cite{ivanova2023running} (e.g. \citet{sainz2023nlp, jacovi2023stop, dong2024generalization, piktus2023roots}) and placing responsibility on the evaluation practitioner. However, it has also been argued that model developers share this responsibility \cite{reuel2024betterbench}, particularly in transparently reporting train-test overlap statistics along with evaluation results \cite{zhang2024language} or planning the evaluation benchmark prior to training \cite{piktus2023roots}. 

\subsection{Evaluation Instance}

System outputs can be judged from various angles, such as correctness or fluency, each corresponding to a distinct evaluation criterion \cite{Zhang_Zhang_Yuan_Liu_Shi_Gui_Zhang_Huang_2024, zhou2022deconstructing}, also referred to as a normative baseline \cite{weidinger2023sociotechnical}. An evaluation criterion refers to a specific aspect of the system's output, necessary to define what constitutes a correct, appropriate, or high-quality response. The choice of criteria depends on both the nature of the task under evaluation \cite{van2019best} and the evaluation target element. Additional criteria may include generalization, reasoning, robustness \cite{hu2024unveiling}, logical coherence, harmlessness \cite{Zhang_Zhang_Yuan_Liu_Shi_Gui_Zhang_Huang_2024}, and naturalness \cite{van2019best}, among others.

If the evaluation criterion is not mathematically defined and no prior annotation of the dataset exists, it typically needs to be applied to the system’s outputs by an evaluator subject, which may be either a human or another AI system. In all cases, it is recommended that evaluation criteria be specified as precisely as possible \cite{zhou2022deconstructing, van2019best} to minimize subjectivity and potential sources of bias.

When an evaluator subject is involved, several additional factors must be considered. These include how the evaluator is selected—for instance, if the evaluator is human, they may be recruited via crowdsourcing platforms; if the evaluator is another LLM, preferably, it should not be the same model under evaluation due to self-preference bias \cite{panickssery2024llm}, also referred to as egocentric bias \cite{li2024leveraging}, and it should not belong to the same model family, as this can introduce bias toward similar linguistic styles \cite{bai2023benchmarking}. Moreover, the evaluator system is typically expected to be more capable than the one being assessed \cite{li2024leveraging}.

Additionally, the evaluator’s level of expertise must be considered. For human evaluators, this last factor may involve distinguishing between laypeople and domain experts; for LLM evaluators, an analogous consideration is whether the model has been fine-tuned specifically for domain-specific evaluation tasks (e.g. \citet{huang2024limitations}).

The nature and instructions of the evaluator’s task must also be specified, such as pairwise comparison -as in Chatbot Arena \cite{chiang2024chatbot}-, rating individual outputs on a Likert scale or assigning a score \cite{li2024leveraging}. The order in which outputs are presented is also important, as both human evaluators \cite{van2019best}, and AI systems are susceptible to position bias \cite{gao2025llm}.  Finally, in human evaluations, it is considered good practice to always report the number of participants along with relevant demographic data (e.g., gender, nationality, age, fluency in the target language, academic background) \cite{weidinger2023sociotechnical} to enhance replicability and allow readers to assess the significance of the results \cite{van2019best}.

Note that in this framework, we distinguish between the concepts of annotators and evaluators. An \textit{annotator}—whether a human or an AI system—is responsible for labeling a test dataset prior to the evaluation of the evaluated subject. This may involve selecting the correct answer in a multiple-choice item, assigning categories to text fragments, or specifying the desired answer to a question. In contrast, an evaluator assesses and annotates the outputs of the evaluated subject, for instance by rating the creativity of a poem generated by an LLM using a Likert scale. While the term 'annotator' is often used in the literature to refer to both roles (e.g. \citet{weidinger2023sociotechnical}), we maintain this distinction for clarity.

\subsection{Measurement}

A metric is a mathematically defined function used to quantify a specific evaluation criterion. Metrics provide a standardized and objective way to measure aspects of system outputs—such as accuracy, coherence, fluency, or similarity to a reference—as they can summarize model performance on a set of tasks and datasets as a single number or score \cite{raji2021ai, eriksson2025can}.

Evaluation practitioners may confuse evaluation criteria with the metrics used to operationalize them, but this confusion risks losing the ability to assess the metric’s appropriateness and distinguish between the metric and the original goal \cite{zhou2022deconstructing}. As Goodhart’s Law states, 'When a measure becomes a target, it ceases to be a good measure.'

Broadly, evaluation metrics can be categorized as either reference-based or reference-free \cite{ito2025reference, li2024leveraging}. Reference-based metrics are used to compare the generated output to a predefined reference, which may be produced manually by human annotators or by another AI system, typically focusing on accuracy, relevance, coherence and similarity to the reference \cite{li2024leveraging}. However, because generating such references can be costly—and in some cases, a ground truth may not be available—reference-free metrics estimate the quality of the output without relying on a reference, often concentrating on its intrinsic qualities, such as fluency or context relevance\footnote{The examples of accuracy, relevance, coherence, fluency and context relevance reflect typical focuses in natural language generation evaluation; other domains may emphasize different intrinsic or comparative qualities.} \cite{li2024leveraging}. Nevertheless, due to certain limitations—such as biases against higher-quality outputs—this kind of metrics are often better suited as diagnostic tools, rather than as definitive measures of task performance \cite{deutsch2022limitations}.

Validity concerns also apply to evaluation metrics. Internal validity may be compromised when inconsistencies in implementation or parameter choices introduce variation that undermines the reliability of metric scores \cite{liao2021are}. External validity issues, in turn, can arise from a mismatch in the evaluation metric of interest \cite{liao2021are, gehrmann2023repairing}. To address these limitations, automatic metrics are often used to complement human evaluation, which has long been considered the gold standard \cite{liao2023rethinking, ruan2024better}. This combination allows practitioners to assess the degree of correlation between automatic and human judgments, and to prioritize which models should undergo more resource-intensive human evaluation processes such as crowdsourcing \cite{zhou2022deconstructing}. For this reason it is also important to justify the choice of automatic metrics \cite{reuel2024betterbench, zhou2022deconstructing}.

Another distinction is between general-purpose metrics and task-specific metrics. General-purpose metrics are widely used across a variety of tasks to assess model performance and can be further categorized following \citet{hu2024unveiling}. Multiple-classification metrics evaluate how effectively an AI system classifies items into multiple groups, examples include Accuracy, Recall, Precision, and F1 score. Token-similarity metrics measure the similarity between texts generated by AI systems and corresponding reference texts; examples include Perplexity, BLEU, ROUGE, BERTScore, and METEOR. Question-answering metrics, such as Strict Accuracy, Lenient Accuracy, and Mean Reciprocal Rank, are used to evaluate performance on QA tasks.

In contrast, task-specific metrics are designed to meet the unique requirements of particular tasks. For example, Translation Edit Rate (TER) can be used to evaluate machine translation output \cite{snover2006study}. Other examples include metrics developed for debate experiments, such as Aggregate Rating and Win Rate \cite{khan2024debating}, as well as the toxicity metric used in red teaming evaluations \cite{hong2024curiositydriven} or Attack Success Rate\cite{lin2025against}.

It is important to report the exact formulas or processes used to calculate these metrics, along with any parameters \cite{reuel2024betterbench} and hyperparameters to allow fair comparisons \cite{gehrmann2022evaluation}.

If the evaluator’s task involves selecting the preferred response from a set, a ranking system is required to aggregate individual judgments and determine which system is preferred overall. In the case of pairwise comparisons, common approaches include the Elo rating system or Points Scoring System \cite{Zhang_Zhang_Yuan_Liu_Shi_Gui_Zhang_Huang_2024}.

Finally, when the goal of an evaluation is to compare the performance of different systems, approaches, or evaluation methods, one or more baselines are typically used as reference points or standards for comparison (see, for example \citet{hong2024curiositydriven, shi2024chops}). Baselines help contextualize the performance of the models under evaluation. By comparing a model’s results against a baseline, researchers can better understand whether improvements are meaningful and how significant they are relative to established expectations. Additionally, human performance in the evaluation may be included as a floor, allowing readers to put the system’s performance into perspective \cite{reuel2024betterbench}.

\subsection{Results Analysis}

Ranking models according to a single quality number is easy and actionable, but, in many circumstances, it is much more important to understand when and why models fail \cite{eriksson2025can}. Therefore, it is also beneficial to conduct a deeper analysis of the evaluation results, as this can reveal valuable insights and guide future research directions \cite{hamalainen-alnajjar-2021-human}. This typically involves providing a concise summary of key findings—both quantitative and qualitative (e.g., analyzing results in relation to dataset complexity levels, if a dataset composition analysis was conducted)—, and outlining potential areas for future work—explaining how the results inform model development and suggest avenues for improvement. Interpreting results in light of contextual thresholds for what constitutes `acceptable' model performance—that may vary across use cases, domains, or applications—is a component of this stage \cite{weidinger2023sociotechnical}.

Additionally, when relevant, statistical analysis should be performed to distinguish genuine performance improvements from random variability. Best practices emphasize treating evaluation processes as structured experiments, applying significance testing and quantifying uncertainty through confidence intervals or resampling techniques \cite{miller2024adding, DBLP:journals/corr/abs-2403-15250}. Methods such as ANOVA, Tukey’s HSD, Student’s t-test, Mann–Whitney U test, and McNemar’s test can be adopted to compare models under various settings, enabling more robust conclusions across different architectures and training regimes \cite{DBLP:journals/corr/abs-2403-15250}. Moreover, careful statistical assessment is useful for detecting biases or artifacts in benchmark performance, ensuring that reported gains reflect true advances rather than contamination or evaluation artifacts (see for example, \citet{zhang2024pacost}). Incorporating these practices helps build more reliable and reproducible evaluations, particularly in high-stakes or competitive settings where small metric differences may lead to significant decisions.

In human evaluations, it is considered good practice to calculate inter-annotator agreement \cite{reuel2024betterbench} as a means of assessing the consistency of the judgments.

Finally, it is recommended to publish system outputs and evaluation code, enabling others to reproduce the findings \cite{eriksson2025can}, re-score outputs under different criteria, or conduct their own statistical analyses \cite{biderman2024lessons}. Facilitating reproducibility is particularly important, as it remains a major concern in this field \cite{reuel2024betterbench}.

\subsection{Transversal Elements of the Conceptual Framework}

\textbf{Evaluation Practitioners.} The human subject responsible for designing and conducting the evaluation may come from industry, government, academia, or act as an independent third-party evaluator. The degree of human involvement can vary depending on the extent to which components of the evaluation are automated \cite{eriksson2025can}. While this may seem self-evident, highlighting this role is important. For example, in human evaluations, the person designing the evaluation should not be the same individual who assesses the system’s outputs, as preexisting hypotheses or expectations may bias their judgment. To mitigate this risk, output assessments are often delegated to external raters, such as those recruited through crowdsourcing platforms.

Additional considerations include the importance of multidisciplinary and diversity. With regard to the former, the natural limitations of the evaluation designers’ knowledge on a potentially infinitely large number of domains and tasks, may lead to generalist approaches that often fail to address the subtle requirements of critical sectors or approaches which could hinder innovation \cite{mcintosh2024inadequacies, eriksson2025can}. As for the latter, evaluation is never neutral \cite{weidinger2023sociotechnical, rauh2024gaps}: many decisions implicitly reflect specific epistemological perspectives regarding how the world is ordered \cite{orr2024ai, eriksson2025can}. Moreover, the social and cultural environments in which evaluations are conducted shape them through shared and often arbitrary assumptions, commitments, and dependencies \cite{eriksson2025can}.

\textbf{Ethical Considerations.} \citet{gupta2024conceptual} present a conceptual framework for balancing information gain and ethical harm in the evaluation of AI systems. They argue that ethical issues arising during evaluation are often overlooked or inadequately predicted, which can diminish the acceptability of the results and the overall value and utility of the information obtained. For example, adversarial testing practices and red teaming may introduce ethically harmful impacts on practitioners or data annotators during the model output labeling process \cite{zhang2024human, gupta2024conceptual, weidinger2023sociotechnical}.

\textbf{Pilot Tests.} Pilot tests refer to preliminary trials conducted prior to the full-scale evaluation to refine the evaluation setup and detect potential problems or vulnerabilities early in the process. This may include verifying the clarity of task instructions and identifying ambiguities or biases in the data or scoring procedures. Pilot testing helps ensure the robustness of the evaluation design and is especially valuable when human evaluators are involved \cite{hamalainen-alnajjar-2021-human} or when novel methods are proposed (see for example, \citet{bai2023benchmarking}).

\textbf{Ablation Studies.} Ablation studies refer to a research methodology used to investigate the contributions of individual components within a system or process. This involves systematically removing or altering specific elements of a system to observe the effect of their absence or modification on the overall performance or behavior. By isolating the impact of each component, researchers can identify which factors are essential for achieving desired outcomes and which are less influential (for examples in evaluations, see \citet{shi2024chops, hong2024curiositydriven}). This approach is particularly useful for refining theories, improving system design, and studying the robustness and efficiency of  complex models or processes \cite{meyes2019ablation}.

\section{Appendix 2: Use Cases}
\label{appendix: 2}
This appendix presents three application cases of our evaluation framework, demonstrating its applicability across a diverse range of existing evaluation settings. The first case involves LLMs acting as evaluators \cite{bai2023benchmarking}, the second focuses on a meta-evaluation of red-teaming methods \cite{lee2025learning}, and the third examines the evaluation with AI agents \cite{rasheed2024large}. While the selection is not exhaustive and some details may be simplified or open to interpretation, the aim is to illustrate how the proposed framework can be applied to analyze real-world evaluation practices.

\begin{longtable}{
  |>{\raggedright\arraybackslash}m{4cm}  
  |>{\raggedright\arraybackslash}m{3.5cm}  
  |>{\raggedright\arraybackslash}m{3.5cm}  
  |>{\raggedright\arraybackslash}m{3.5cm}| 
}
\hline
\textbf{Elements of the Conceptual Framework} &
\textbf{Paper 1: \textit{``Benchmarking Foundation Models with Language-Model-as-an-Examiner''}} &
\textbf{Paper 2: \textit{``Learning Diverse Attacks on Large Language Models for Robust Red Teaming and Safety Tuning''}} &
\textbf{Paper 3: \textit{``Large Language Model Evaluation Via Multi
AI Agents: Preliminary Results''}} \\
\hline
\endfirsthead
\multicolumn{4}{|c|}{\textit{Continued from previous page}} \\
\hline
\textbf{Elements of the Conceptual Framework} &
\textbf{Paper 1: \textit{``Benchmarking Foundation Models with Language-Model-as-an-Examiner''}} &
\textbf{Paper 2: \textit{``Learning Diverse Attacks on Large Language Models for Robust Red Teaming and Safety Tuning''}} &
\textbf{Paper 3: \textit{``Large Language Model Evaluation Via Multi
AI Agents: Preliminary Results''}} \\
\hline
\endhead

\multicolumn{4}{|c|}{\textbf{Evaluation Target}} \\
\hline

\textbf{Capability} & Examiners: the ability of LLMs to serve as examiners and their potential biases in a decentralized setting. Evaluated LLMs: question answering. & Capacity to elicit undesirable responses from a target LLM, through the generation of diverse and effective attacks. & Code generation. \\
\hline

\textbf{Objective} & To evaluate if the proposed decentralized LLM-based evaluation method mitigate the lack of diversity and scope of the generated questions and bias during evaluation compared to a centralized evaluation. & To evaluate whether the proposed method for improving the diversity and effectiveness of attacks performs better, worse, or comparably to existing approaches. & To evaluate and compare the code generation capabilities of various LLMs using a novel multi-agent AI framework with automatic verification. \\
\hline

\multicolumn{4}{|c|}{\textbf{Task}} \\
\hline

\textbf{Task Mode}& Open-ended question answering. & Attacker Model Task: Prompt generation. Target Model Task: Open-ended question answering. & Open-ended code generation. \\
\hline

\textbf{Steps}& Single-step & Single-Step & Multi-step \\
\hline

\textbf{Interactions} & Both single-turn and multi-turn. & Single-turn. & Single-turn interaction (however, the authors reported several rounds of interaction for the OpenAI series). \\
\hline

\multicolumn{4}{|c|}{\textbf{Evaluated Subject}} \\
\hline

\textbf{Evaluated Subject} & For the centralized evaluation  8 LLMs were evaluated: BLOOMZ, Flan-T5, Flan-UL2, GLM- 130B, LLaMA, Vicuna-13B, and ChatGPT. For the descentralized evaluation 4 LLMs were evaluated: ChatGPT, Claude, Vicuna-13B, Bard. & Since this is a meta-evaluation, the primary subject of evaluation is the red-teaming method itself, specifically the attacker LLM. The paper mentions the evaluation of two attacker models: GPT-2 small and Llama-3.2-1B. To validate the method’s effectiveness, five target LLMs were red-teamed: GPT-2, Gemma-2b-it, Dolly-v2-7b, Llama-2-7b-chat, and Llama-3.1-8B-Instruct. & 7 LLMs were evaluated: ChatGPT-4 Turbo, GPT-4, GPT-3.5 Turbo, GPT-3.5, Google Bard, Llama and Hugging Face (CodeBERT). \\
\hline

\textbf{Identity Conditions} & Number of parameters and training procedure were reported for the centralized evaluation: BLOOMZ (the 176B model), Flan-T5 (the XXL model, 11B), Flan-UL2 (20B), GLM- 130B, LLaMA (the 13B model and the 65B model), Vicuna-13B, and ChatGPT. These models are categorized based on their training procedure: whether they have undergone Supervised Fine-Tuning (SFT) or not. The first 6 models are trained without SFT, whereas the last 2 models are fine-tuned. & The proposed method use GFlowNet fine-tuning, followed by a secondary smoothing phase, to train the attacker model to generate diverse and effective attack prompts. The attacker model is defined by its probabilistic formulation—specifying its objective, sampling policy, training constraints, and fine-tuning procedure —as detailed in Section 3.2 of the paper. The attackers models include GPT-2 Small (124M parameters) and Llama-3.2-1B (1.2B parameters)// The target models evaluated include GPT-2 (124M parameters),  Gemma-2b-it (2B parameters), Dolly-v2-7b (7B parameters),  Llama-2-7b-chat (7B parameters), and Llama-3.1-8B-Instruct (8B parameters). & Number of parameters: ChatGPT-4 Turbo - 1.96 trillion, GPT-4 - 1.76 trillion, GPT-3.5 Turbo - 154 billion, GPT-3.5 - 125 billion, Google Bard - 1.56 trillion, Llama - 70 billion and Hugging Face (CodeBERT) - 355M. \\
\hline

\textbf{Operational Context} & Centralized evaluation: the temperature was set to 0 for both the examiner and the subject models to ensure reproducibility. & All target models are accessed in a black-box setting, meaning only prompt-response behavior is observed. & LLMs were accessed directly via raw APIs. 7 AI Agents were responsible for interacting with a different evaluated LLM to retrieve programming code.\\
\hline

\multicolumn{4}{|c|}{\textbf{System Inputs}} \\
\hline

\textbf{Amount} & Centralized evaluation: Single-Round: 10,000 questions. Multi-round: randomly selected 1000 questions. // Descentralized evaluation: each examiner model posed 100 questions. & Not reported. & 10 input prompts used across the 7 evaluated models. \\
\hline

\textbf{Modality}& Text. & Text. & Text. \\
\hline

\textbf{Input Source} & The test dataset, namely LMExamQA, was generated by an LLM. To ensure wide coverage of knowledge, Google Trends Categories were chosen as the domain taxonomy, and n domains from it were randomly selected. For each domain, the LLM was prompted to generate m distinct questions. Their designed prompt is formulated to ensure that the generated questions possess three essential characteristics: diversified question forms, varied cognitive levels, and most importantly, assurance that the LM has a comprehensive understanding of the knowledge surrounding the question it poses. The resulting dataset was analyzed according to Bloom’s taxonomy. & Adversarial prompts generated by a separate attacker model. & Each LLM was provided with identical, high-level natural language descriptions of programming tasks. The test dataset was created ad-hoc, aiming at incorporating diverse coding tasks to comprehensively evaluate the model's coding capabilities. \\
\hline

\textbf{Prompt Techniques} & Centralized Evaluation: For models without SFT, their assess their 0-shot and 5-shot performance. & Not Reported. & The prompting technique used in the paper can be characterized as standardized zero-shot instruction prompting. The natural language descriptions of programming tasks were provided without examples or contextual demonstrations. \\
\hline

\multicolumn{4}{|c|}{\textbf{Evaluation Instance}} \\
\hline

\textbf{Evaluation Criteria }& (1) Accuracy. This assesses the extent to which the provided response accurately answers the question. (2) Coherence. This evaluates the logical structure and organization of the response and the degree to which it can be comprehended by non-specialists. (3) Factuality. This examines whether the response contains factual inaccuracies. (4) Comprehensiveness. This gauges whether the response encompasses multiple facets of the question, thus providing a thorough answer. & Diversity and Toxicity. & Syntactic correctness, adherence to the prompt, computational efficiency in terms of execution and resource usage, code accuracy (functional correctness of the generated code as per the given description) and innovation demonstrated in the solutions. \\
\hline

\textbf{Evaluator Subject }&  The novel decentralized method incorporates multiple models to serve as examiners, namely Peer-examination. Centralized experiment: GPT-4. Descentralized experiment: GPT-4, Claude (Claude-instant), ChatGPT, Bard, and Vicuna-13B. To justify the reliability of the LM examiner, it was tasked with generating ground-truth answers, and a random sample of 100 questions was evaluated by human experts. & \textbf{RoBERTa} hate speech classifier (Vidgen et al., 2021) for GPT-2 and dolly-v2-7b, and \textbf{Llama-Guard} (LLM) for Llama-2-7b-chat, Llama-3.1- 8B-Instruct, and Gemma-2b-it. & The primary evaluator subject in this study is an automated verification agent, specifically designed to assess code generated by multiple LLMs. This agent employs the HumanEval benchmark and computes the pass@k metric (with k=1), allowing for consistent, objective, and reproducible evaluation of functional correctness across models. \\
\hline

\textbf{Evaluator Task} & Likert Scale Scoring and Ranking. & The evaluator's task is to assess the toxicity of the output generated by the target language models in response to adversarial prompts. Given a prompt–response pair (x,y), the evaluator—either the RoBERTa hate speech classifier or Llama-Guard—assigns a toxicity score or label, which determines whether the response is considered harmful. The output is  binary variable denoting toxicity. A prompt is considered  toxic if the toxicity classifier assigns a score greater than 0.5. & Rating of individual outputs via the HumanEval Benchmark.  Additionally, a quality rating, depicted with stars in section 4.2 of the paper provides a subjective assessment of the code based on criteria such as readability, efficiency, and adherence to best practices. \\
\hline

\multicolumn{4}{|c|}{\textbf{Measurement}} \\
\hline

\textbf{Metric} & Likert scale scoring and a variant of pairwise comparison, namely ranking. Liket Scale: Each criteria is scored on a scale of 1 to 3, ranging from worst to best. The evaluator was also asked to provide an overall score ranging from 1 to 5, based on the scores assigned to the previous 4 criteria. This score serves as an indicator of the overall quality of the answer. In the pairwise comparison evaluators are given two responses and are tasked with determining which is superior, taking into account the four evaluation criteria. An evaluation of the metrics was conducted to know whether they correlate with human judgments. & \textbf{Toxicity: Toxicity rate} which is the percentage of generated prompts that are toxic. \textbf{Diversity: Average pairwise cosine distance} to measure diverity.  A model named MiniLMv2 (sentence-transformer model) is used to embed the generated prompts so the distance can be calculated. & The primary metric used in this study is pass@k, from the HumanEval benchmark that quantifies functional correctness by measuring the proportion of generated code outputs that successfully pass predefined unit tests on the first attempt. In the experiments, a pass@1 approach was utilized, meaning that we evaluated the models based on their first attempt at generating a solution. This approach was chosen to reflect a more realistic scenario where a developer would use the model’s first output. \\
\hline

\textbf{Baseline} & The centralized evaluation method was used as a baseline to compare against the proposed approach. & The method was compaired against some relevant red-teaming baselines: \textbf{Supervised Fine-tuning (SFT), In-Context Learning (ICL), REINFORCE, PPO + Novelty, GFlowNet, GFlowNet + MLE}. & Not used. \\
\hline

\multicolumn{4}{|c|}{\textbf{Result Analysis}} \\
\hline

\textbf{Qualitative Analysis}& They used a win-rate heatmap to visualize the results. Conclusions: 1. The results adhere to the scaling law of LMs; 2. Few-shot leads to more substantial improvement on higher cognitive-level questions; 3. SFT primarily plays a crucial role in aligning LM’s responses for task adaptation, rather than enriching the model’s knowledge — especially in the context of higher-level questions that demand more sophisticated answers; 4. LLMs can provide factually correct and coherent responses, but struggle for more comprehensive accurate answers; 5. They observe that excluding ChatGPT and Vicuna-13B, all examinee models exhibit a notable decrease in performance in the second round. This suggests that while these models initially demonstrated a robust understanding and knowledge base, their performance deteriorated when faced with more complicated questions. & The proposed method, GFlowNet + MLE, outperforms all baseline approaches in generating diverse and effective adversarial prompts across five target language models. It achieves a better balance between toxicity rate and prompt diversity, compared to prior methods like PPO + Novelty and REINFORCE, which either collapse to a few toxic prompts or fail to generate effective ones. Prompts generated by GFlowNet + MLE also show strong transferability, successfully attacking unseen models such as GPT-4o and larger LLaMA variants. Moreover, safety-tuning models using these prompts leads to more robust defenses against other red-teaming attacks without harming general performance. The second stage (MLE smoothing) is also computationally efficient, adding robustness with minimal additional training cost. & Conclusions: 1. Model Performance Is Not Strictly Correlated with Parameter Size. Despite having significantly fewer parameters than models like GPT-4 or Google Bard, GPT-3.5 Turbo outperformed all other models in terms of functional correctness; 2. Code Quality Varies Widely Across Models. The subjective star-based ratings revealed notable differences in code quality beyond correctness; 3. Instruction Adherence Is Uneven Across Systems. While all models received identical natural language prompts, their ability to faithfully implement the described functionality varied; 4. Evaluation Framework Enhances Comparative Interpretability. The multi-agent evaluation setup, combined with the verification agent and HumanEval benchmark, proved effective in revealing relative strengths and weaknesses across LLMs. \\
\hline

\textbf{Statistical Analysis }& Not reported. & There is descriptive statistical analysis (means and standard deviations across multiple runs), but there's no inferential statistical testing like p-values or confidence intervals. & No Statistical Analysis was mentioned. Results were reported in terms of raw accuracy counts (i.e., number of correct code generations out of 10 per model) and qualitative star-based ratings. \\
\hline

\textbf{Future Work }& Expanding the framework to incorporate more domain-specific language models, or even vision language models, could potentially offer a more holistic evaluation. & The approach is not limited to text tokens and future work can explore the applicability to red-team multimodal models (e.g., text-to-image models ). Further, an interesting area of future work is extending the approach to the jailbreaking setting, where an attacker language model generates a suffix for an adversarial query prompt. Finally, in addition to red-teaming, it would be interesting to apply our method to generate prompts which can improve model performance on different tasks. & The future work outlined in the paper involves three key directions: (1) expanding the evaluation dataset from 10 to 50 input descriptions to enable broader and more robust analysis; (2) integrating the MBPP (Massively Multitask Benchmark for Python) to complement HumanEval and provide more diverse and challenging test cases; and (3) conducting real-world validation by sharing the model with 20 practitioners from diverse backgrounds, with the aim of collecting qualitative feedback to enhance the model’s practical relevance and usability. \\
\hline

\end{longtable}


\end{document}